\documentclass{article}

\usepackage[nonatbib,final]{nips_2016}

\usepackage[utf8]{inputenc} 
\usepackage[T1]{fontenc}    
\usepackage{url}            
\usepackage{booktabs}       
\usepackage{amsfonts}       
\usepackage{nicefrac}       
\usepackage{microtype}      

\usepackage{color}
\usepackage{graphicx}
\usepackage{amssymb,amsfonts,amsmath,amsthm}
\usepackage{paralist}
\usepackage{mathtools}
\usepackage{appendix}
\usepackage[all,dvips,arc,curve,color,frame]{xy}
\usepackage{algorithm}
\usepackage[noend]{algpseudocode}
\usepackage{bbm}
\usepackage{tikz}
\usetikzlibrary{shapes.gates.logic.US,trees,positioning,arrows,automata,decorations.pathmorphing,chains}

\def \By {\mathbf{y}}

\def \aset {\mathcal{A}}

\newcommand{\argmin}{\operatornamewithlimits{argmin}}

\newtheorem{thm}{Theorem}[section]

\title{Resource Constrained Structured Prediction}

\author{
  Tolga Bolukbasi\textsuperscript{1}, Kai-Wei Chang\textsuperscript{2}, Joseph Wang\textsuperscript{1}, Venkatesh Saligrama\textsuperscript{1}\\
  \textsuperscript{1} Boston University, Boston, MA\\
  \textsuperscript{2} Microsoft Research, Cambridge, MA\\
}

\begin{document}

\maketitle

\begin{abstract}
We study the problem of structured prediction under test-time budget constraints. We propose a novel approach applicable to a wide range of structured prediction problems in computer vision and natural language processing. Our approach seeks to adaptively generate computationally costly features during test-time in order to reduce the computational cost of prediction while maintaining prediction performance. We show that training the adaptive feature generation system can be reduced to a series of structured learning problems, resulting in efficient training using existing structured learning algorithms. This framework provides theoretical justification for several existing heuristic approaches found in literature. We evaluate our proposed adaptive system on two structured prediction tasks, optical character recognition (OCR) and dependency parsing and show strong performance in reduction of the feature costs without degrading accuracy.
\end{abstract}

\section{Introduction}
\label{sec:intro}
Structured prediction is a powerful and flexible framework for making a joint prediction over mutually dependent output variables. It has been successfully applied to a wide range of computer vision and natural language processing tasks ranging from text classification to human detection. However, the superior performance and flexibility of structured predictors come at the cost of computational complexity. In order to construct computationally efficient algorithms, a trade-off must be made between the expressiveness and speed of structured models.

The cost of inference in structured prediction can be broken down into three parts: acquiring the features, evaluating the part responses and solving a combinatorial optimization problem to make a prediction based on part responses. Past research has focused on evaluating part responses and solving the combinatorial optimization problem, and proposed efficient inference algorithms for specific structures (e.g., Viterbi and CKY parsing algorithms) and general structures~(e.g., variational inference~\cite{jordan1999introduction}). 
However, these methods overlook feature acquisition and part response, which are bottlenecks when the underlying structure is relative simple or is efficiently solved. 

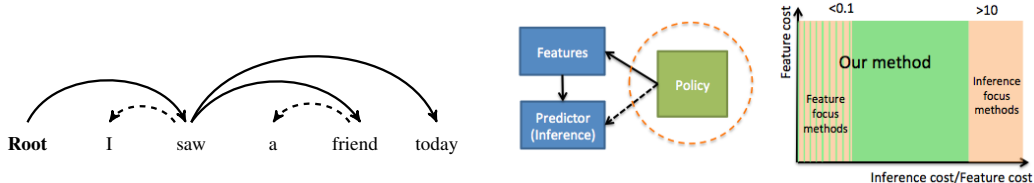
\begin{figure*}
\vspace{-1cm}
 \begin{minipage}[!b]{0.45\linewidth}
 \vfill
 \resizebox{\textwidth}{!}{
   \begin{tikzpicture}[start chain,->,>=stealth',node
    distance=0.5cm,very thick,every node/.style={anchor=base}]
    \node   (A) at (0,1)   {\bf  Root};
    \node   (B) at (1.5,1) {I};
    \node   (C) at (3,1) {saw};
    \node   (D) at (4.5,1)  {a};
    \node   (E) at (6,1)  {friend};
    \node   (F) at (7.5,1)  {today};
    \path (0,1.5)   edge [bend left=64]   (2.9,1.5)
          (3,1.5)   edge [bend left=55]   (6.1,1.5)
          (3,1.5) edge [bend left=66]   (7.5, 1.5)
          (2.7, 1.5) edge[bend right=55  draw, dashed]   (1.5, 1.5)
          (5.9, 1.5) edge[bend right=55 draw, dashed ]  (4.5, 1.5);
    \end{tikzpicture}
  }
  \end{minipage}\hfill
  \begin{minipage}{0.52\linewidth}
  	\includegraphics[width=\linewidth]{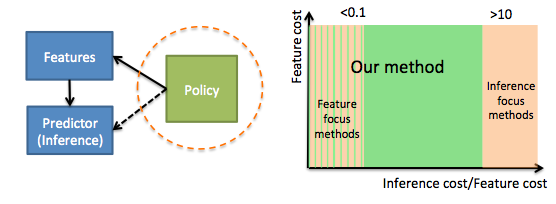}
  \end{minipage} 
  \vspace{-.4cm}
\caption{\footnotesize{\textbf{Left:} When predicting the dependency tree, some dependencies (e.g., the dashed edges) are easily resolved, and there is less need for expressive features in making a prediction. \textbf{Right:} Our system diagram and operating region. When relatively feature and inference costs are both not insignificant the policy must carefully balance the overhead costs due to feedback from the predictor with feature acquisition costs.}
}
\label{fig:motivate}
\vspace{-.4cm}
\end{figure*}

Consider the dependency parsing task, where the goal is to create a directed tree that describes semantic relations between words in a sentence. The task can be formulated as a structured prediction problem, where the inference problem concerns finding the maximum spanning trees (MSTs) in a directed graphs \cite{mcdonald2005non}. Each node in the graph represents a word, and the directed edge $(x_i, x_j)$ represents how likely $x_j$ depends on $x_i$. Fig. \ref{fig:motivate} shows an example of a dependency parse and the trade off between a rich set of features and the  prediction time. Introducing complex features has the potential to increase system performance, however they only distinguish among a small subset of ``difficult'' parts.  Therefore, computing complex features for all parts on every example is both computationally costly and unnecessary to achieve high levels of performance. %

We address the problem of structured prediction under test-time budget constraints, where the  goal is to learn a system that is computationally efficient during test-time with little loss of predictive performance. We consider test-time costs associated with the computational cost of evaluating feature transforms or acquiring new sensor measurements. 
Intuitively, our goal is to learn a system that identifies the parts in each example incorrectly classified/localized using ``cheap'' features and additionally yield large reductions in error over the entire structure given ``expensive'' features due to improved distinguishability and relationships to other parts in the example.

We consider two forms of the budgeted structured learning problem, prediction under expected budget constraints and anytime prediction. For both cases, we consider the streaming test-time scenario where the system operates on each test example without observation or interaction of other test examples.
In the {\it expected budget constraint} setting, the system chooses features to acquire for each example, to minimize prediction error subject to an average feature cost constraint. A fixed budget is given by the user during training time, and during test-time, the system is tasked with both allocating resources as well as determining the features to be acquired with the allocated resources.
In the {\it anytime structured prediction} setting, the system chooses features to be acquired sequentially for each example to minimize prediction error at each time step, allowing for accurate prediction at anytime. No budget is specified by the user during training time. Instead,the system sequentially chooses features to minimize prediction error at any time as features are acquired. This setting requires a single system able to predict for any budget constraint on each example.

We learn systems capable of adaptive feature acquisition for both settings. We propose learning policy functions to exploit relationships between parts and adapt to varying length examples. This problem naturally reduces policy learning to a structured learning problem, allowing the original model to be used with minor modification. The resulting systems reduce prediction cost during test-time with minimal loss of predictive performance.
We summarize our contributions as follows:\\ \textbullet \, Formulation of structured prediction under expected budget constraints and anytime prediction.\\ \textbullet  \, Reduction of both these settings to conventional structured prediction problems.\\ \textbullet  \, Demonstration that structured models benefit from having access to features of multiple complexities and can perform well when a only a subset of parts use the expensive features.

\section{Budgeted Structured Learning}
We begin with reviewing structured prediction problem and formulating it under an expected budget constraint.
We then extend the formulation to anytime structured prediction.

\textbf{Structured Prediction:} %
The goal in  structured prediction is to learn a function, $F$, that maps from an input space, $\mathcal{X}$, to a structure space, $\mathcal{Y}$. In contrast to multi-class classification,
the space of outputs $\mathcal{Y}$ is not simply categorical
but instead is assumed to be some exponential space of outputs (often of varying size dependent on the feature space) 
containing some underlying structure, generally represented by multiple parts and relationships between parts.
For example, in dependency parsing, $x\in \mathcal{X}$ are features representing a sentence (e.g., words, pos tags), and $y\in \mathcal{Y}$ is a parse tree. 

In a structured prediction model, 
the mapping function $F$ is often modeled as $F\equiv \max_{y\in Y}\Psi(x,y)$, where
$\Psi: (X,Y) \rightarrow R$ is a scoring function. We assume the score can be broken up into sub-scores across components $C$, $\Psi(x,y) = \sum_{c\in C} \psi_c(x, y_c)$, where $y_c$ is the output assignment associated with the component $c$. The number of sub-components, $|C|$, varies across examples.
For the dependency parsing example, 
each $c$ is an edge in the directed graphs, and $y_c$ is an indicator variable for whether the edge is in the parse tree. The score of a parse tree consists of the scores $\psi_c(x,y_c)$ of all its edges.

\subsection{Structured Prediction Under an Expected Budget}\label{section.expected_budget}

Our goal is to reduce the cost of prediction during test-time (representing computational time, energy consumption, etc.). We consider the case where a variety of scoring functions are available to be used for each component. Additionally, associated with each scoring function is an evaluation cost (such as the time or energy consumption required to extract the features for the scoring function).%

For each example, we define a state $S \in  \mathcal{S}$, where the space of states is defined $ \mathcal{S}=\{0,1\}^{K \times |C|}$, representing which of the $K$ features is used for the $|C|$ components during prediction. In the state, the element $S(k,c)=1$ indicates that the $k^{\text{th}}$ feature will be used during prediction for  component $c$. For any state $S$, we define the evaluation cost:
$
c(S)=\sum\nolimits_{c \in C}\sum\nolimits_{k \in K}S(k,c) \delta_{k},
$
where $\delta_k$ is the (known) cost of evaluating the $k^{th}$ feature for a single part.

We assume that we are given a structured prediction model $F:\mathcal{X}\times \mathcal{S}\rightarrow \mathcal{Y}$ that maps from a set of features $X \in \mathcal{X}$ and a state $S \in \mathcal{S}$ to a structured label prediction $\hat{Y} \in \mathcal{Y}$. For a predicted label, we have a loss $L:\mathcal{Y}\times \mathcal{Y} \rightarrow \mathbbm{R}$ that maps from a predicted and true structured label, $\hat{Y}$ and $Y$, respectively, to an error cost, generally either an indicator error, $L(\hat{Y},Y)=1-\prod_{i=1}^{k}\mathbbm{1}_{\hat{Y}(i)=Y(i)}$, or a Hamming error, $L(\hat{Y},Y)=\sum_{i=1}^{k}\mathbbm{1}_{\hat{Y}(i)\neq Y(i)}$. For an example $(X,Y)$ and state $S$, we now define the modified loss
$
C(X,Y,S)=L\left(F(X,S),Y\right)+\lambda c(S)
$
that represents the error induced by predicting a label from $X$ using the sensors in $S$ combined with the cost of acquiring the sensors in $S$, where $\lambda$ is a trade-off pattern adjusted according to the budget $B$\footnote{Our framework does not restrict the type of modified loss, $C(X,Y,S)$, or the state cost, $C(S)$ and extends to general losses}. A small value of $\lambda$ encourages correct classification at the expense of feature cost, whereas a large value of $\lambda$ penalizes use of costly features, enforcing a tighter budget constraint.
We define a policy $\pi:\mathcal{X}\rightarrow \mathcal{S}$ that maps from the feature space $\mathcal{X}$ and the initial state $S_0$ to a new state. For ease of reference, we refer to this policy as the feature selection policy. Our goal is to learn a policy $\pi$ chosen from a family of functions $\Pi$ that, for a given example $X$, maps to a state with minimal expected modified loss,
$\pi^*=\argmin\nolimits_{\pi \in \Pi} \mathbbm{E}_{\mathcal{D}}\left[C\left(X,Y,\pi\left(X\right)\right)\right].$
In practice, $\mathcal{D}$ denotes a set of I.I.D training examples:
\begin{align}\label{eq.policy_erm_singlestep}
\pi^*=\argmin\nolimits_{\pi \in \Pi} \sum\nolimits_{i=1}^{n}C\left(X_i,Y_i,\pi\left(X_i\right)\right).
\end{align}
Note that the objective of the minimization can be expanded with respect to the space of states, allowing the optimization problem in \eqref{eq.policy_erm_singlestep} to be expressed
$\pi^*=\argmin\nolimits_{\pi \in \Pi} \sum\nolimits_{i=1}^{n}\sum\nolimits_{S \in\mathcal{S}}C\left(X_i,Y_i,S\right)\mathbbm{1}_{\pi\left(X_i\right)=S}.$
From this, we can reformulate the problem of learning a policy as a structured learning problem.
\begin{thm}\label{thm.onestep_theorem}
The minimization in \eqref{eq.policy_erm_singlestep} is equivalent to the structured learning problem:
{\small
\begin{align*}
\argmin_{\pi \in \Pi} \sum_{i=1}^{n}&\sum_{S \in \mathcal{S}}\big(\max_{\tilde{S}\in \mathcal{S}}C(X_i,Y_i,\tilde{S})- C\left(X_i,Y_i,S\right)\big)\mathbbm{1}_{\pi\left(X_i\right)\neq S}.
\end{align*}}
\end{thm}
Proofs can be found in Suppl. Material. Theorem \ref{thm.onestep_theorem} maps the policy learning problem in \eqref{eq.policy_erm_singlestep} to a weighted structured learning problem. For each example $X_i$, an example/label pair is created for each state $S$ with an importance weight representing the savings lost by not choosing the state $S$.

Unfortunately, the expansion of the cost function over the space of states introduces the summation over the combinatorial space of states. To avoid this, we instead introduce an approximation to the objective in \eqref{eq.policy_erm_singlestep}. Using a single indicator function, we formulate the approximate policy
\begin{align}\label{eq.erm_onestep_approx}
\hat{\pi}=\argmin\nolimits_{\pi \in \Pi} \sum\nolimits_{i=1}^{n} \Bigg[W(X_i,Y_i)\mathbbm{1}_{\pi\left(X_i\right)\neq S^*(X_i,Y_i)}
+ C\left(X_i,Y_i,S^*(X_i,Y_i)\right)\Bigg],
\end{align}
\begin{flalign}
	\label{eq:S_star}
\mbox{where the pseudo-label is defined: } & \quad S^*(X_i,Y_i)=\argmin\nolimits_{S \in \mathcal{S}} C\left(X_i,Y_i,S\right)&&
\end{flalign}
and the example weight is defined as
$W(X_i,Y_i)=\max_{S'\in \mathcal{S}}C\left(X_i,Y_i,S'\right)-C\left(X_i,Y_i,S^*(X_i,Y_i)\right)$.

This formulation reduces the objective from a summation over the combinatorial set of states to a single indicator function for each example and represents an upper-bound on the original risk.
\begin{thm}\label{thm.onestep_ub}
The objective in \eqref{eq.erm_onestep_approx} is an upper-bound on the objective in \eqref{eq.policy_erm_singlestep}.
\end{thm}

Note that the second term \eqref{eq.erm_onestep_approx} is not dependent on 
$\pi$. Thus, Theorem \ref{thm.onestep_ub} leads to an efficient algorithm for learning a policy function $\pi$ by solving an importance-weighted structured learning problem:
\begin{align}\label{eq.onestep_ub_alg}
\hat{\pi}=\argmin\nolimits_{\pi \in \Pi} \sum\nolimits_{i=1}^{n}&W(X_i,Y_i)\mathbbm{1}_{\pi\left(X_i\right)\neq S^*(X_i,Y_i)},
\end{align}
where each example $X_i$ having a pseudo-label $S^*(X_i,Y_i)$ and importance weight $W(X_i,Y_i)$.

\noindent
{\bf Combinatorial Search Space:}
Finding the psuedo-label in Eqn. \eqref{eq:S_star} involves searching over the combinatorially large search space of states, $\mathcal{S}$, which is computationally intractable. Instead, we present 
trajectory-based and parsimonious pseudo-labels for approximating $S^*$.

\noindent
{\it Trajectory Search:}
The trajectory-based pseudo-label is a greedy approximation to the optimization in Eqn. \eqref{eq:S_star}. %
To this end, define ${\cal S}_i^+$ as the 1-stage feasible transitions: 
$
{\cal S}_i^t = \{S\mid d(S,{\hat S}^{t-1}_i) \leq 1,\,S \wedge {\hat S}^{t-1}_i = {\hat S}^{t-1}_i \},
$
where $d$ is the Hamming distance. %
We define a trajectory of states $\hat{S}^t_i$ %
where 
$\hat{S}^t_i=\argmin_{S \in {\cal S}_i^t} C(X_i,Y_i,S)$.
The initial state is assumed to be $\hat{S}^{0}_i=0^{K \times |C|}$ where none of the $K$ features are evaluated for the $C$ components.
For each example $i$, we obtain a trajectory $S^0_i,S^1_i,\ldots, S^T_i$, where the terminal state $S^T_i$ is the all-one state.
We choose the pseudo-label from the trajectory:
$\hat{S}^*_i=\argmin\nolimits_{S \in \{\hat{S}^{0}_i,\ldots,\hat{S}^T_i\}} C(X_i,Y_i,S).$
Note that by restricting the search space of states differing by a single component, the approximation needs to only perform a polynomial search over states as opposed to the exhaustive combinatorial search in Eqn. \eqref{eq:S_star}. Observe that the modified loss is not strictly decreasing, as the cost of adding features may outweigh the reduction in loss at any time. %
Empirically, this approach is computationally tractable and is shown to produce strong results.

\noindent
{\it Parsimonious Search:}
Rather than a trajectory search, which requires an inference update as we acquire more features, we consider an alternative one stage update here. The idea is to look for 1-step transitions that can potentially improve the cost. We then simultaneously update all the features that produce improvement. This obviates the need for a trajectory search. 
In addition we can incorporate a guaranteed loss improvement for our parsimonious search. %
${\cal S}_i^+ \in \argmin\nolimits_{S \in {\cal S}^t_i} \mathbf{1}_{\{C(X_i,Y_i,S^{t-1}_i)\geq C(X_i,Y_i,S) + \tau \}}.$
Note that the potential candidate transitions can be non-unique and thus we generate a collection of potential state transitions, ${\cal S}_i^+$.
To obtain the final state we take the union over these transitions, namely,
$
{\hat S}^* = \bigvee_{S \in {\cal S}_i^+} S.
$
Suppose we set the margin $\tau=0$, replace the cost-function with the loss function then this optimization is relatively simple (assuming that acquiring more features does not increase the loss). This is because the new state is simply the collection of transitions where the sub-components are incorrect. Finding the parsinomious pseudo-label is computationally efficient and empirically shows similar performance to the trajectory-based pseudo-label.

Choosing the pseudo-label requires knowledge of the budget $B$ to set the cost trade-off parameter $\lambda$. If the budget is unspecified or varies over time, a system capable of adapting to changing budget demands is necessary. To handle this scenario, we propose an anytime system in the next section.

\subsection{Anytime Structured Prediction}\label{sec.anytime_pol}
In many applications, the budget constraint is unknown a priori or varies from example to example due to changing resource availability and an expected budget system as in Section \ref{section.expected_budget} does not yield a feasible system. We instead consider the problem of learning an anytime system \cite{grubb2012speedboost}. In this setting, a single system is designed such that when a new example arrives during test-time, features are acquired until an arbitrary budget constraint (that may vary over different examples) is met for the particular example. Note that an anytime system is a special case of the expected budget constrained system. Instead of an expected budget, instead a hard per-example budget is given. A single system is applied to all feasible budgets, as opposed to learning unique systems for each budget constraint.

We model the anytime learning problem as sequential state selection. The goal is to select a trajectory of states, starting from an initial state $S_0=0^{k\times |C|}$ where all components use features with negligible cost. To select this trajectory of states, we define policy functions $\pi_1,\ldots,\pi_T$, where $\pi_t:\mathcal{X}\times \mathcal{S}\rightarrow \mathcal{S}$ is a function that maps from a set of structured features $X$ and current state $S$ to a new state $S'$.

The sequential selection system is then defined by the policy functions $\pi_1,\ldots,\pi_{T}$. For an example $X$, the policy functions produce a trajectory of states $S^1(X)\ldots,S^{T}(X)$ defined as follows:
$S^t(X)=\pi^t(X,S^{t-1}(X)), \qquad S^0(X)=S^0.$

Our goal is to learn a system with small expected loss at any time $t \in [0,T]$. Formally, we define this as the average modified loss of the system over the trajectory of states:
\begin{align}\label{eq.exp_anytime_policy}
\pi_1^*,...,&\pi_T^*=\argmin\nolimits_{\pi_1,...,\pi_T\in\Pi}({1}/{T})\mathbbm{E}_{\mathcal{D}}\left[\sum\nolimits_{t=1}^{T}C\left(X,Y,S^t(X)\right)\right]
\end{align}
where $\Pi$ is a user-specified family of functions. Unfortunately, the problem of learning the policy functions is highly coupled due to the dependence of the state trajectory on all policy functions. 
Note that as there is no fixed budget, the choice of $\lambda$ dictates the behavior of the anytime system. Decreasing $\lambda$ leads to larger increases in classification performance at the expense of budget granularity. %

We propose a greedy approximation to the policy learning problem by sequentially learning policy functions $\pi_1,\ldots,\pi_T$ that minimize the modified loss:
\begin{align}\label{eq.exp_greedy_policy}
\pi_t=\argmin\nolimits_{\pi \in \Pi} \mathbbm{E}_{\mathcal{D}}\left[C\left(X,Y,S^t(X)\right)\right]
\end{align}
for $t \in \{1,\ldots,T\}$. Note that the $\pi_t$ selected in \eqref{eq.exp_greedy_policy} does not take into account the future effect on the loss in \eqref{eq.exp_anytime_policy}. We consider $\pi_t$ in \eqref{eq.exp_greedy_policy} to be a greedy approximation as it is instead chosen to minimize the immediate loss at time $t$.

We restrict the output space of states for the policy $\pi_t$ to have the same non-zero components as the previous state with a single feature added. This space of states can be defined 
$
\mathcal{S}(S)=\{S'|d(S',S)=1,S'\wedge S=S\},
$
where $d$ is the Hamming distance. Note that this mirrors the trajectory used for the trajectory-based pseudo-label.

As in Section \ref{section.expected_budget}, we take an empirical risk minimization approach to learning policies. To this end we sequentially learn a set of function $\pi_1,\ldots,\pi_T$ minimizing the risk:
{\small
\begin{align}\label{eq.erm_greedy_ind}
\argmin_{\pi \in \Pi}\sum _{i=1}^{n}\smashoperator[r]{\sum_{s \in \mathcal{S}(S^{t-1}(X_i))}}&C\left(X_i,Y_i,s\right)\mathbbm{1}_{\pi(X_i,S^{t-1}(X_i))=s},
\end{align}}
enumerating over the space of states that the policy $\pi_t$ can map each example. Note that the space of states $\mathcal{S}(S^{t-1}(X_i))$  may be empty if all features are acquired for example $X_i$ by step $t-1$.

\begin{figure}[!t]
\centering
\begin{minipage}[t]{0.47\textwidth}
\begin{algorithm}[H]
	 \caption{Anytime Policy Learning\label{alg:policy_training}}
	 \begin{algorithmic}
	 \State {\bf input} Training set, $\{X_i, Y_i\}_{i=1,...,n}$ 
	 \State {\bf set} $S_i^0 = \mathbf{0} \: \forall i = 1,...,n, t=1$
	 \While{$\mathcal{A}(S_i)\neq\emptyset$ for any $i$}
         \State Train $\pi_t$ according to Thm. \ref{thm.structured_learning_eq} 
	     \For{$i \in [n]$}
		     \State Update states: $S_i^{t}=\pi_t(X_i,S_i^{t-1})$
	     \EndFor
         \State $t\gets t+1$
	\EndWhile
	\State {\bf return} $\pi=\{\pi_1,\ldots,\pi_T\}$
	 \end{algorithmic}
\end{algorithm}
\end{minipage}
 \hfill
\begin{minipage}[t]{0.47\textwidth}
 \begin{algorithm}[H]
	 \caption{Anytime Structured Prediction \label{alg:test-time_policy}}
	 \begin{algorithmic}
	 \State {\bf input} Policy, $\pi_1,...,\pi_T$, Example, $X$, Budget, $B$
     \State {\bf set} $S^0 = \mathbf{0}, t=0$
	 \While{$\aset(S^t) \neq \emptyset$ and $c(S^t)<B$}
	     \State $S^{t+1}=\pi_t(X,S^t)$, $t \gets t+1$
	\EndWhile
	\State {\bf return} $\By =F(X, S^t)$
	 \end{algorithmic}
 \end{algorithm}
 \end{minipage}
 \vspace{-.4cm}
 \end{figure}

As in Thm. \ref{thm.onestep_theorem}, the problem of learning the sequence of policy functions $\pi_1,\ldots,\pi_T$ can be viewed as a weighted structured learning problem.
\begin{thm}\label{thm.structured_learning_eq}
The optimization problem in \eqref{eq.erm_greedy_ind} is equivalent to solving an importance weighted structured learning problem using an indicator risk of the form:
\begin{align}\label{eq.thm_sl_eq}
 \argmin_{\pi \in \Pi}\sum_{i=1}^{n}\smashoperator[r]{\sum_{s \in \mathcal{S}(S^{t-1}(X_i))}}\,\,\,W(X_i,Y_i,s)\mathbbm{1}_{\pi(X_i,S^{t-1}(X_i))\neq s},
\end{align}
\begin{flalign*}
\mbox{where the weight is defined: } & W(X_i,Y_i,s)=\max\nolimits_{s'\in \mathcal{S}(S^{t-1}(X_i))}\,\,\,C(X_i,Y_i,s')- C\left(X_i,Y_i,s\right).&&
\end{flalign*}
This is equivalent to an importance weighted structured learning problem, where each state $s$ in $\mathcal{S}(S^{t-1}(X_i))$ defines a pseudo-label for the example $X_i$ with an associated importance $\left(\max_{s'\in \mathcal{S}(S^{t-1}(X_i))}C(X_i,Y_i,s')- C\left(X_i,Y_i,s\right)\right)$.
\end{thm}

Theorem \ref{thm.structured_learning_eq} reduces the problem of learning a policy to an importance weighted structured learning problem. Replacement of the indicators with upper-bounding convex surrogate functions results in a convex minimization problem to learn the policies $\pi_1,...,\pi_T$. In particular, use of a hinge-loss surrogate converts this problem to the commonly used structural SVM. Experimental results show significant cost savings by applying this sequential policy.

The training algorithm is presented in Algorithm \ref{alg:policy_training}. At time $t=0$, the policy $\pi_1$ is trained to minimize the immediate loss. Given this policy, the states of examples at time $t=1$ are fixed, and $\pi_2$ is trained to minimize the immediate loss given these states. The algorithm continues learning policies until every feature for every example as been acquired. During test-time, the system sequentially applies the trained policy functions until the specified budget is reaches, as shown in Algorithm \ref{alg:test-time_policy}.

\section{Related Work}
\label{sec:related}

 Multi-class prediction with test-time budget has received significant attention (see e.g., ~\cite{viola2001robust, chen2012classifier, busa2012fast, karayev2013dynamic, xu2013cost, trapeznikov2013supervised, kusner2014feature, wang2014model, wang2014lp}). Fundamentally, multi-class classification based approaches cannot be directly applied to structured settings for two reasons: (1) \underline{Structured Feature Selection Policy:} Unlike multi-class prediction, in a structured setting, we have many parts with associated features and costs for each part. This often requires a coupled adaptive part by part feature selection policy applied to varying structures; (2) \underline{Structured Inference Costs}: In contrast to multi-class prediction, structured prediction requires solving a constrained optimization problem in test-time, which is often computationally expensive and must be taken into account.

Strubell et al. \cite{strubell2015learning} improve the speed of a parser that operates on search-based structured prediction models, where joint prediction is decomposed to a sequence of decisions.
In such a case, resource-constrained multi-class approaches can be applied, however this reduction only applies to search-based models that are fundamentally different from the graph-based models we
discussed (with different types of theoretical guarantees and use cases). Applying their policy to the case of graphical models requires repeated inferences, dramatically increasing the computational cost when inference is slow.\footnote{The equivalent policy of \cite{strubell2015learning} applied to our inference algorithm is marked as the myopic policy in our experiments. Due to the high cost of repeated inference, the resulting policy is computationally intensive.}

Similar observations apply to Weiss et al. 
\cite{weiss2013dynamic,weiss2013learning}, who present a scheme for adaptive feature selection assuming the computational costs of policy execution and inference are negligible. Their approach uses a reinforcement learning scheme, requiring inference at each step of their policy to estimate rewards. For complex inference tasks, repeatedly executing the policy (performing 
inference) can negate any computational gains induced 
by adaptive feature selection (see Fig. 3 in \cite{weiss2013learning}).

He et al. \cite{he2013dynamic} use imitation learning to adaptively select features for dependency parsing. Their approach can be viewed as an approximation of Eqn. \eqref{eq.onestep_ub_alg} with a parsimonious search. Although their policy avoids performing inference to estimate reward, multiple inferences are required for each instance due to the design of action space. Overhead is avoided by exploiting the specific inference structure (maximal spanning tree over fully connected graph), and it is unclear if it can be generalized.

Methods to increase the speed of inference (predicting the given part responses) have been proposed \cite{weiss2010structured,shi2015learning}. These approaches can be incorporated into our approach to further reduce computational cost and therefore are complementary. More focused research has improved the speed of individual algorithms such as object detection using deformable parts models 
\cite{felzenszwalb2010object,zhu2014active} and dependency parsing~\cite{he2013dynamic,strubell2015learning}. These methods are specialized, failing to generalize to varying graph size and/or structures and relying on problem-specific heuristics or algorithm-specific properties.

Adaptive features approaches have been designed to improve accuracy, 
including easy-first decoding strategies~\cite{goldberg2010efficient,stoyanov2012easy}, however these methods focus on performance as opposed to computational cost.

\section{Experiments\label{sec:experiments}}

In this section, we demonstrate the effectiveness of the proposed algorithm on 
two structured prediction tasks in different domains: dependency parsing and OCR.
We report the results on both anytime and expected case policies and refer to the latter one as one-shot policy.
Our focus is mainly on the policy and not on achieving the state of the art performance on either of these domains. 

At a high-level, policies for resource constrained structured prediction must manage \& tradeoff benefits of three resources, namely, feature acquisition costs, intermediate inferencing costs, and policy overhead costs that decides between feature acquisition and inferencing. Some methods as described earlier account for feature costs but not inference and overhead costs. Other methods incorporate inference into their policy (meta-features) for selecting new features but do not account for the resulting policy overhead. Our approach poses policy optimization as a structured learning problem and in turn jointly optimizes these resources as demonstrated empirically in our experiments.

We compare our system to the Q-learning approach in \cite{weiss2013learning} and two baselines: a uniform policy and a myopic policy. 
The uniform policy takes random part level actions. 
The uniform policy will help us show that the performance of our policy does not come from removing redundant features, but clever allocation of them among samples.
As a second baseline, we adapt the myopic policy used by \cite{trapeznikov2013supervised} to the structured prediction case.  The myopic policy runs the structured predictor initially on all cheap features, then looks at the total confidence of the classifier normalized by the sample size (e.g. sentence length). If the confidence is below a threshold, it chooses to acquire expensive features for all positions.
Finally, we compare against the Q-learning method proposed by \cite{weiss2013learning}. This method requires global features for structures with varying size. From now on we will refer to features that require access to more than one part as complex features and part level features as simple features. In their case, they use confidence feedback from the structured predictor which induces additional inference overhead for the policy. In addition to this, it is not straightforward to apply this approach to do part by part feature selection on structures with varying sizes.

\begin{figure}[!t]
\centering
\begin{minipage}{.5\textwidth}
\centering
\includegraphics[width=1.1\textwidth]{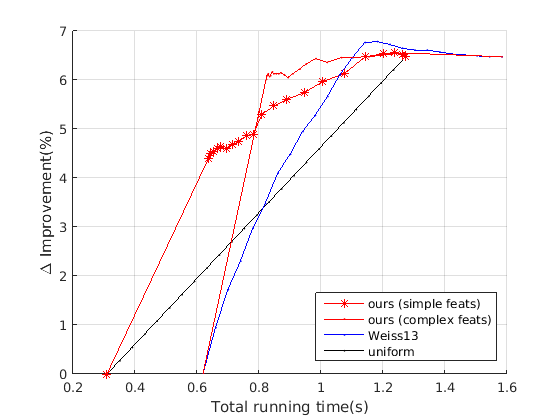}
\end{minipage}\hfill
\begin{minipage}{.4\textwidth}
\centering
\includegraphics[width=\textwidth]{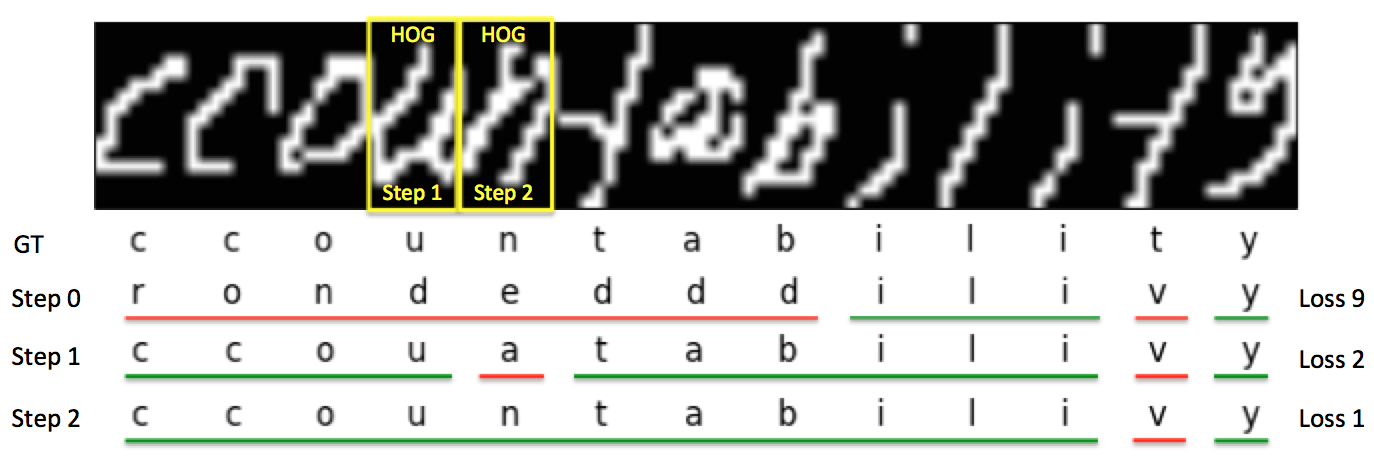}
\caption{\footnotesize{An example word from the OCR test dataset is shown. Note that the word is initially incorrectly identified due to degradation in letters "u" and "n". The letter classification accuracy  increases after the policy acquires the HOG features at strategic positions.}}\label{fig:ocr_easy_example}
\end{minipage}
\caption{The performance of our one-shot policy (in red) is compared to the uniform strategy (in black) and policy of Weiss et. al. \cite{weiss2013learning} for the OCR dataset. Although the policy with complex features is more efficient with features, the simple feature policy has a lower total run-time in the low budget region due to the overhead of additional inference.}\label{fig:ocr_results}
\vspace{-0.5cm}
\end{figure}

We adopt Structured-SVM~\cite{tsochantaridis2005large} to solve the policy learning problems for expected and anytime cases defined in \eqref{eq.onestep_ub_alg} and \eqref{eq.thm_sl_eq}, respectively. 
For the structure of the policy $\pi$ we use a graph with no edges due to its simplicity. In this form, the policy learning problem can be written as a sample weighted SVM. We discuss the details in the appendix due to space constraints.

We show in the following that complex features indeed benefit the policy, but simple features perform better for cases where the inference time and feature costs are comparable and the additional overhead is unwanted. Finally, we show that part by part selection outperforms global selection.

\noindent
{\bf Optical Character Recognition} We tested our algorithm on a sequence-label problem, the OCR dataset \cite{taskar2003max} composed of 6,877 handwritten words, where each word is represented as a sequence of 16x8 binary letter images.  We use a linear-chain Markov model, and similar to the setup in \cite{weiss2013dynamic, wang2014model}, use raw pixel values and HOG features with 3x3 cell size as our feature templates.  We split the data such that 90\% percent is used for training and 10\% is used for test.

Fig. \ref{fig:ocr_results} shows the average letter accuracy vs. total running time. The proposed system reduces the budget for all performance levels, with a savings of up to 50 percent at the top performance. Note that Weiss13 can not operate on part by part level when the graph structure is varying. We see that using complex part by part selection has significant advantage over using uniform feature templates. Furthermore, Fig \ref{fig:ocr_easy_example} shows the behavior of the policy on an individual example for the anytime model, significant gains in accuracy are made in first several steps by correctly identifying the noisy letters.

\noindent
{\bf Dependency Parsing}
We follow the setting in \cite{he2013dynamic} and conduct experiments on English Penn Treebank (PTB) 
corpus~\cite{marcus1993building}. 
All algorithms are implemented based on the graph-based dependency 
parser~\cite{mcdonald2005non} in Illinois-SL 
library~\cite{chang2015illinoissl},
where the code is optimized for speed. 
Two sets of feature templates are considered for the parser.  \footnote{
Complex features often contribute to small performance improvement.
Adding complex or redundant features can easily yield arbitrarily large speedups, and comparing speedups of different systems with different accuracy levels is not meaningful (see Fig. 3 in \cite{he2013dynamic}). In addition, greedy-style parser such as \cite{strubell2015learning} might be 
faster by nature. Discussing different architecture and features is outside the scope of this paper. 
}
The first ($\psi^{\text{Full}}$) considers the part-of-speech (POS) tags and lexicons
of $x_i$, $x_j$, and their surrounding words (see \cite{mcdonald2005non}). The other ($\psi^{\text{POS}}$) only considers the POS features. 
The policy assigns one of these two feature templates to each word in the sentence,
such that all the directed edges $(x_i, x_j)$ corresponding to the word $x_i$ share the same feature templates.
The first feature template, $\psi^{\text{POS}}$, takes 165 $\mu$s per word and the second feature template, $\psi^{\text{Full}}$, takes 275 $\mu$s per word to extract the features and compute edge scores. The decoding by Chu–Liu-Edmonds algorithm is 75 $\mu$s per word, supporting our hypothesis that feature extraction makes a significant portion of the total running time yet the inference time is not negligible.
Due to the space limit, we present further details of the experiment setting in the appendix.
\begin{figure}[!t]
\centering
\begin{minipage}{0.53\textwidth}
\centering
\includegraphics[width=\textwidth]{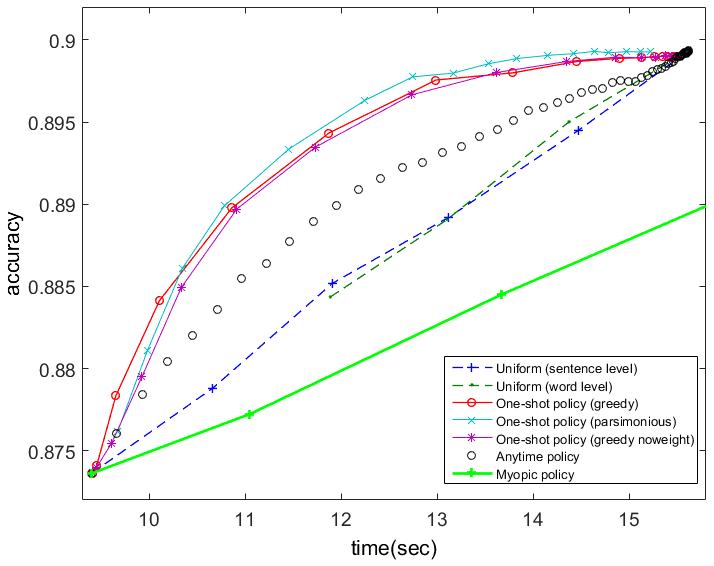}
\end{minipage}\hfill
\begin{minipage}{0.43\textwidth}
\parbox{0.9\linewidth}{\label{fig.dep_parse_example}\includegraphics[width=\textwidth,height=0.47\textwidth]{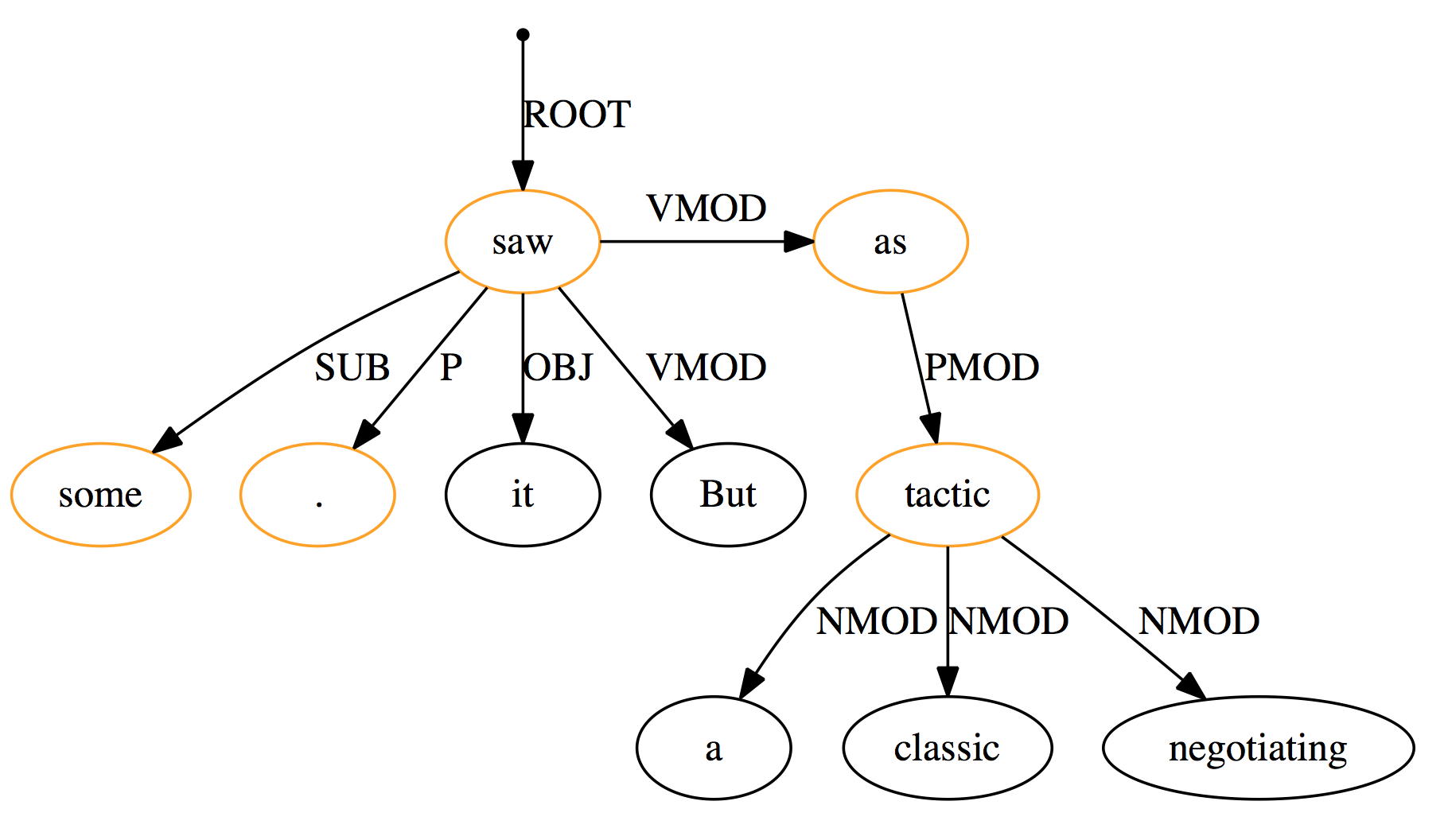}}
\parbox{0.9\linewidth}{\label{fig.depth_dis}\includegraphics[width=\textwidth]{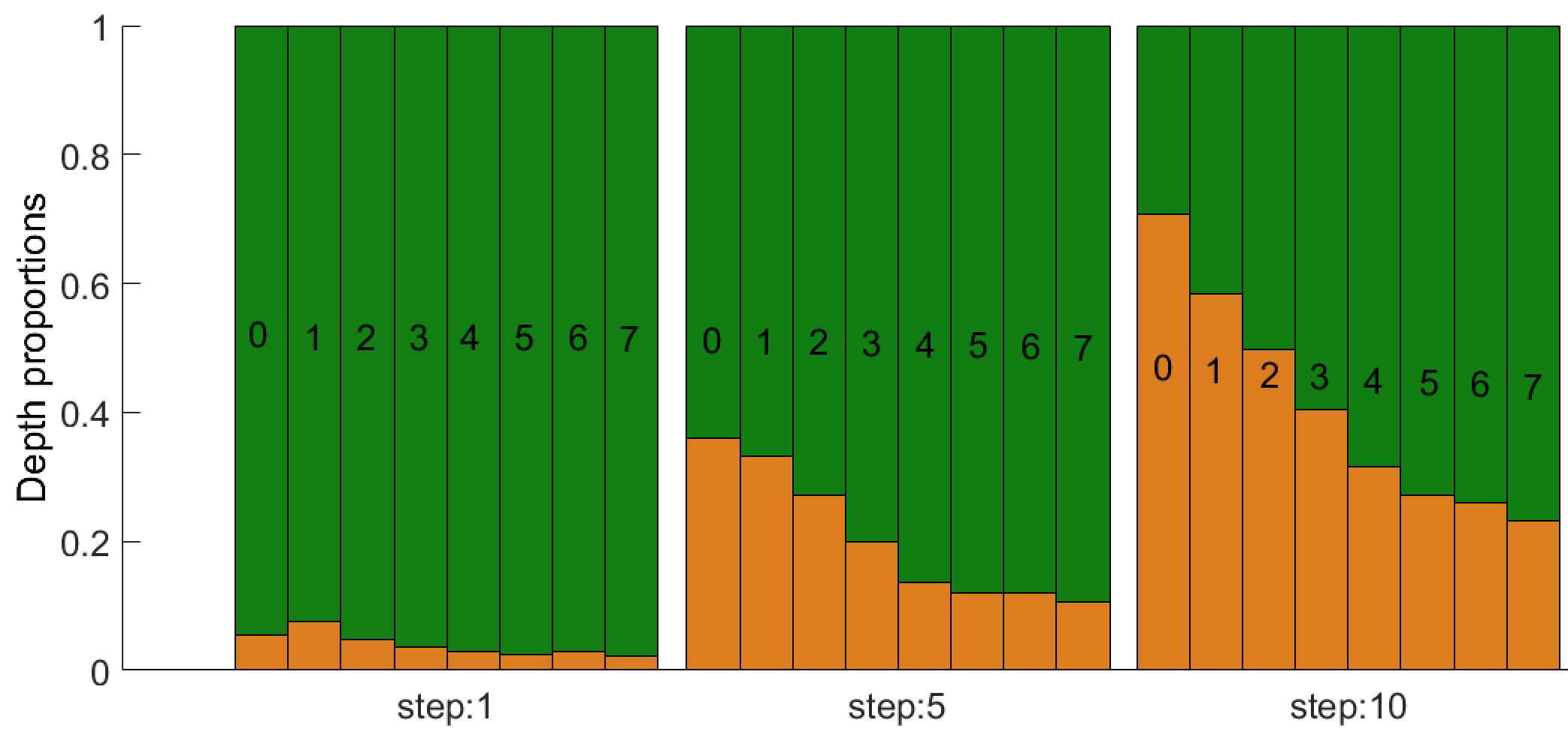}}
\end{minipage}
\caption{\footnotesize{\textbf{Left:} \footnotesize{Performance of various adaptive policies for varying budget levels (dependency tree accuracy vs. total execution time), is compared to a uniform strategy on word and sentence level, and myopic policy for the 23 section of PTB dataset.}\label{fig:depparse_results} \textbf{Right:} Distribution of parse-tree depth for words that use cheap (green) or expensive features (orange) for anytime policy. Time increases from left to right. Each group of columns show the distribution of depths from 0(root) to 7. The policy is concentrated on acquiring features for lower depth words. A sentence example also shows this effect. It is easy to identify parents of the adjectives and determiner. However, additional features(orange) are required for the root(verb), subject and object.}}
\end{figure}

Fig. \ref{fig:depparse_results} shows the test performance (unlabeled 
attachment accuracy) along with inference time.
We see that all one-shot policies 
perform similarly, losing negligible accuracy when using half of the 
available expensive features. %
When we apply the length dictionary filtering heuristic in \cite{he2013dynamic,rush12Vine},
our parser achieves 89.7\% UAS on PTB section 23 with overall running 
time merely 7.5 seconds (I/O excluded, 10s with I/O) and obtains 2.9X total 
speed-up with losing only 1\% UAS comparing to the baseline.  
\footnote{This heuristic only works for parsing. Therefore, we exclude it when presenting Figure \ref{fig:depparse_results} as it does not reflect the performance of policies in general.}
This significant speed-up over an efficient implementation is remarkable. 
\footnote{In contrast, the baseline system in  \cite{he2013dynamic} is slow than us by 
about three times. When operating in the accuracy level of 90\%, 
Figure 3 in \cite{he2013dynamic} shows that their final system takes about 
20s. 
We acknowledge that \cite{he2013dynamic} use different features, policy settings, and hardware 
from ours; therefore these numbers might not be comparable.}

Although marginal, one-shot policy with greedy trajectory has the strongest performance in low budget regions. This is because the greedy trajectory search has better granularity than parsimonious search in choosing positions that decrease the loss early on. The anytime policy is below one-shot policy for all budget levels. As discussed in \ref{sec.anytime_pol}, the anytime policy is more constrained in that it has to achieve a fixed budget for all examples. The naive myopic policy performs worse than uniform since it has to run inference on samples with low confidence two times, adding approximately 4.5 seconds of extra time for the full test dataset. %
We then explore the effect of importance weights for the greedy policy. We notice a small improvement. We hypothesize that this is due to the policy functional complexity being a limiting factor.

We also conduct ablative studies to better understand the policy behavior.
Fig. \ref{fig.depth_dis} shows the distribution of depth for the words that use expensive and cheap features in the ground truth dependency tree. We expect investing more time on the low-depth words (root in the extreme) to yield higher accuracy gains. We observe this phenomenon empirically, as the policy concentrates on extracting features close to the root. 
%


\clearpage
\appendix
\section{Proofs}
\paragraph{Proof of Theorem 2.1}\mbox{}\\
The objective in (1) can be expressed:
\begin{align*}
\sum_{i=1}^{n}&\sum_{S \in \mathcal{S}}C\left(X_i,Y_i,S\right)\mathbbm{1}_{\pi(X_i)=S}\\
=\sum_{i=1}^{n}&\sum_{S \in \mathcal{S}}C\left(X_i,Y_i,S\right)\left(1-\mathbbm{1}_{\pi(X_i)\neq S}\right)\\
=\sum_{i=1}^{n}&\Bigg[\max_{S'\in \mathcal{S}}C(X_i,Y_i,S')-\max_{S'\in \mathcal{S}}C(X_i,Y_i,S')\\
&+\sum_{S \in \mathcal{S}}C\left(X_i,Y_i,S\right)\left(1-\mathbbm{1}_{\pi(X_i)\neq s}\right)\Bigg].
\end{align*}
Note that $\sum_{S \in \mathcal{S}}\left(1-\mathbbm{1}_{\pi(X_i)\neq S}\right)=1$, allowing for further simplification:
\small{
\begin{align*}
=\sum_{i=1}^{n}&\Bigg[\max_{S'\in \mathcal{S}}C(X_i,Y_i,s')\\
&+\sum_{s \in \mathcal{S}}\left(C\left(X_i,Y_i,S\right)-\max_{S'\in \mathcal{S}}C(X_i,Y_i,S')\right)\\
&\left(1-\mathbbm{1}_{\pi(X_i)\neq S}\right)\Bigg]\\
=\sum_{i=1}^{n}&\Bigg[\max_{S'\in \mathcal{S}}C(X_i,Y_i,S')\\
&+\sum_{S \in \mathcal{S}}\left(\max_{S'\in \mathcal{S}}C(X_i,Y_i,S')-C\left(X_i,Y_i,S\right)\right)\left(\mathbbm{1}_{\pi(X_i)\neq s}-1\right)\Bigg].
\end{align*}}
Removing constant terms (that do not affect the output of the $\argmin$) yields the expression in Thm. 2.1.

\paragraph{Proof of Theorem 2.2}\mbox{}\\
For a single example/label pair $(X_i,Y_i)$, consider the two possible values of the term in the summation of (4). In the event that $\pi(X_i)=S^*(X_i,Y_i)$:
\begin{align*}
W(X_i,Y_i)&\mathbbm{1}_{\pi(X_i)\neq S^*(X_i,Y_i)}+C(X_i,Y_i,S^*(X_i,Y_i))\\
&=C(X_i,Y_i,S^*(X_i,Y_i)),
\end{align*}
which is equivalent to the value of (2) if $\pi(X_i)=S^*(X_i,Y_i)$. Otherwise, $\pi(X_i)\neq S^*(X_i,Y_i)$, and therefore:
\begin{align*}
W(X_i,Y_i)&\mathbbm{1}_{\pi(X_i)\neq S^*(X_i,Y_i)}+C(X_i,Y_i,S^*(X_i,Y_i))\\
&=W(X_i,Y_i)+ C(X_i,Y_i,S^*(X_i,Y_i))\\
&=\max_{S'\in \mathcal{S}}C(X_i,Y_i,S').
\end{align*}
This is an upper-bound on (1), and therefore (2) is a valid upper-bound on (1).

\paragraph{Proof of Theorem 2.3}\mbox{}\\
Note that the objective in (7) can be expressed:
\begin{align*}
\sum_{i=1}^{n}&\sum_{S \in \mathcal{S}(S^{t-1}(X_i))}C\left(X_i,Y_i,S\right)\mathbbm{1}_{\pi(X_i,S^{t-1}(X_i))=S}\\
=\sum_{i=1}^{n}&\sum_{S \in \mathcal{S}(S^{t-1}(X_i))}C\left(X_i,Y_i,S\right)\left(1-\mathbbm{1}_{\pi(X_i,S^{t-1}(X_i))\neq S}\right)\\
=\sum_{i=1}^{n}&\Bigg[\max_{S'\in \mathcal{S}(S^{t-1}(X_i)}C(X_i,Y_i,S')\\
-&\max_{S'\in \mathcal{S}(S^{t-1}(X_i))}C(X_i,Y_i,S')\\
&+\sum_{S \in \mathcal{S}(S^{t-1}(X_i))}C\left(X_i,Y_i,S\right)\left(1-\mathbbm{1}_{\pi(X_i,S^{t-1}(X_i))\neq S}\right)\Bigg].
\end{align*}
Note that $\sum_{S \in \mathcal{S}(S^{t-1})(X_i)}\left(1-\mathbbm{1}_{\pi(X_i,S_{t-1}(X_i))\neq S}\right)=1$, allowing for further simplification:
\begin{align*}
=\sum_{i=1}^{n}&\Bigg[\max_{S'\in \mathcal{S}(S^{t-1})(X_i)}C(X_i,Y_i,S')\\
&+\sum_{S \in \mathcal{S}(S^{t-1}(X_i))}\Bigg(C\left(X_i,Y_i,S\right)\\
-&\max_{S'\in \mathcal{S}(S^{t-1}(X_i))}C(X_i,Y_i,S')\Bigg)\left(1-\mathbbm{1}_{\pi(X_i,S^{t-1}(X_i))\neq s}\right)\Bigg]\\
=\sum_{i=1}^{n}&\Bigg[\max_{S'\in \mathcal{S}(S^{t-1}(X_i))}C(X_i,Y_i,S')\\
&+\sum_{S \in \mathcal{S}(S^{t-1}(X_i))}\Bigg(\max_{S'\in \mathcal{S}(S^{t-1})(X_i))}C(X_i,Y_i,S')\\
-&C\left(X_i,Y_i,S\right)\Bigg)\left(\mathbbm{1}_{\pi(X_i,S^{t-1}(X_i))\neq S}-1\right)\Bigg].
\end{align*}
Removing constant terms (that do not affect the output of the $\argmin$) yields the expression in (8).

\section{Implementation details}
\paragraph{Dependency parsing}
We split PTB corpus into two parts, Sections 02-22 and Section 23 as training and test sets.  
We then conduct a modified cross-validation mechanism to train the feature selector and the dependency parser.
Note that the cost of policy is dependent on the structured predictor. Therefore, learning policy on the same training set of the predictor may cause the structured loss to be overly optimistic. We follow the cross validation scheme in 
to deal with this issue by splitting the training data into $n$ folds. For each fold, we generate label predictions based on the structured predictor trained on the remaining folds. Finally, we gather these label predictions and train a policy on the complete data.

The dependency parser is trained by the averaged Structured Perceptron model
with learning rate and number of epochs set to be 0.1  and 50, respectively. 
This setting achieves the best test performance as reported in 
Notice that if we trained two dependency models with different feature sets separately the scale of the edge scores may be different, resulting sub-optimal test performance.
To fix this issue, we generate data with random edge features and train the model to minimize the joint loss over all states.
\begin{align*}
min_w \lambda||w||^2 + \frac{1}{n}\sum_{i=1}^{n} E_{S} [L(F_w(X_i,S),Y_i)]
\end{align*}

Finally, We found that for dependency parsing expensive features are only necessary in several critical locations in the sentence. Therefore, budget levels above 10\% turned out to be unachievable for any feature-tradeoff parameter lambda in the pseudo-labels. To obtain those regions, we varied the class weights of feature templates in the training of one-shot feature selector.

\bibliographystyle{abbrv}
{\small
\bibliography{budgeted_sl}  

\begin{thebibliography}{10}

\bibitem{busa2012fast}
R.~Busa-Fekete, D.~Benbouzid, and B.~K{\'{e}}gl.
\newblock {Fast classification using sparse decision DAGs}.
\newblock In {\em 29th International Conference on Machine Learning (ICML
  2012)}, pages 951--958. Omnipress, 2012.

\bibitem{chang2015illinoissl}
K.-W. Chang, S.~Upadhyay, M.-W. Chang, V.~Srikumar, and D.~Roth.
\newblock {IllinoisSL: A JAVA Library for Structured Prediction}.
\newblock {\em arXiv preprint arXiv:1509.07179}, 2015.

\bibitem{chen2012classifier}
M.~Chen, K.~Q. Weinberger, O.~Chapelle, D.~Kedem, and Z.~Xu.
\newblock {Classifier cascade for minimizing feature evaluation cost}.
\newblock In {\em International Conference on Artificial Intelligence and
  Statistics}, pages 218--226, 2012.

\bibitem{felzenszwalb2010object}
P.~F. Felzenszwalb, R.~B. Girshick, D.~McAllester, and D.~Ramanan.
\newblock {Object detection with discriminatively trained part-based models}.
\newblock {\em Pattern Analysis and Machine Intelligence, IEEE Transactions
  on}, 32(9):1627--1645, 2010.

\bibitem{goldberg2010efficient}
Y.~Goldberg and M.~Elhadad.
\newblock {An efficient algorithm for easy-first non-directional dependency
  parsing}.
\newblock In {\em Human Language Technologies: The 2010 Annual Conference of
  the North American Chapter of the Association for Computational Linguistics},
  pages 742--750. Association for Computational Linguistics, 2010.

\bibitem{grubb2012speedboost}
A.~Grubb and D.~Bagnell.
\newblock {Speedboost: Anytime prediction with uniform near-optimality}.
\newblock In {\em International Conference on Artificial Intelligence and
  Statistics}, pages 458--466, 2012.

\bibitem{he2013dynamic}
H.~He, H.~{Daum{\'{e}} III}, and J.~Eisner.
\newblock {Dynamic Feature Selection for Dependency Parsing.}
\newblock In {\em EMNLP}, pages 1455--1464, 2013.

\bibitem{jordan1999introduction}
M.~I. Jordan, Z.~Ghahramani, T.~S. Jaakkola, and L.~K. Saul.
\newblock {An introduction to variational methods for graphical models}.
\newblock {\em Machine learning}, 37(2):183--233, 1999.

\bibitem{karayev2013dynamic}
S.~Karayev, M.~Fritz, and T.~Darrell.
\newblock {Dynamic feature selection for classification on a budget}.
\newblock In {\em International Conference on Machine Learning (ICML): Workshop
  on Prediction with Sequential Models}, 2013.

\bibitem{kusner2014feature}
M.~J. Kusner, W.~Chen, Q.~Zhou, Z.~E. Xu, K.~Q. Weinberger, and Y.~Chen.
\newblock {Feature-Cost Sensitive Learning with Submodular Trees of
  Classifiers.}
\newblock In {\em AAAI}, pages 1939--1945, 2014.

\bibitem{marcus1993building}
M.~P. Marcus, M.~A. Marcinkiewicz, and B.~Santorini.
\newblock {Building a large annotated corpus of English: The Penn Treebank}.
\newblock {\em Computational linguistics}, 19(2):313--330, 1993.

\bibitem{mcdonald2005non}
R.~McDonald, F.~Pereira, K.~Ribarov, and J.~Haji{\v{c}}.
\newblock {Non-projective dependency parsing using spanning tree algorithms}.
\newblock In {\em Proceedings of the conference on Human Language Technology
  and Empirical Methods in Natural Language Processing}, pages 523--530.
  Association for Computational Linguistics, 2005.

\bibitem{rush12Vine}
A.~Rush and S.~Petrov.
\newblock Vine pruning for efficient multi-pass dependency parsing learned
  prioritization for trading off accuracy and speed.
\newblock In {\em NAACL}, 2012.

\bibitem{shi2015learning}
T.~Shi, J.~Steinhardt, and P.~Liang.
\newblock {Learning Where to Sample in Structured Prediction.}
\newblock In {\em AISTATS}, 2015.

\bibitem{stoyanov2012easy}
V.~Stoyanov and J.~Eisner.
\newblock {Easy-first Coreference Resolution.}
\newblock In {\em COLING}, pages 2519--2534. Citeseer, 2012.

\bibitem{strubell2015learning}
E.~Strubell, L.~Vilnis, K.~Silverstein, and A.~McCallum.
\newblock {Learning Dynamic Feature Selection for Fast Sequential Prediction}.
\newblock {\em arXiv preprint arXiv:1505.06169}, 2015.

\bibitem{taskar2003max}
B.~Taskar, C.~Guestrin, and D.~Koller.
\newblock {Max-Margin Markov Networks}.
\newblock In {\em Advances in Neural Information Processing Systems}, page
  None, 2003.

\bibitem{trapeznikov2013supervised}
K.~Trapeznikov and V.~Saligrama.
\newblock {Supervised sequential classification under budget constraints}.
\newblock In {\em Proceedings of the Sixteenth International Conference on
  Artificial Intelligence and Statistics}, pages 581--589, 2013.

\bibitem{tsochantaridis2005large}
I.~Tsochantaridis, T.~Joachims, T.~Hofmann, and Y.~Altun.
\newblock {Large margin methods for structured and interdependent output
  variables}.
\newblock In {\em Journal of Machine Learning Research}, pages 1453--1484,
  2005.

\bibitem{viola2001robust}
P.~Viola and M.~Jones.
\newblock {Robust real-time object detection}.
\newblock {\em International Journal of Computer Vision}, 4, 2001.

\bibitem{wang2014model}
J.~Wang, T.~Bolukbasi, K.~Trapeznikov, and V.~Saligrama.
\newblock {Model selection by linear programming}.
\newblock In {\em Computer Vision--ECCV 2014}, pages 647--662. Springer, 2014.

\bibitem{wang2014lp}
J.~Wang, K.~Trapeznikov, and V.~Saligrama.
\newblock {An LP for Sequential Learning Under Budgets}.
\newblock In {\em Proceedings of the Seventeenth International Conference on
  Artificial Intelligence and Statistics}, pages 987--995, 2014.

\bibitem{weiss2013dynamic}
D.~Weiss, B.~Sapp, and B.~Taskar.
\newblock {Dynamic structured model selection}.
\newblock In {\em Proceedings of the IEEE International Conference on Computer
  Vision}, pages 2656--2663, 2013.

\bibitem{weiss2010structured}
D.~Weiss and B.~Taskar.
\newblock {Structured Prediction Cascades}.
\newblock In {\em International Conference on Artificial Intelligence and
  Statistics}, pages 916--923, 2010.

\bibitem{weiss2013learning}
D.~J. Weiss and B.~Taskar.
\newblock Learning adaptive value of information for structured prediction.
\newblock In {\em Advances in Neural Information Processing Systems}, pages
  953--961, 2013.

\bibitem{xu2013cost}
Z.~Xu, M.~Kusner, M.~Chen, and K.~Q. Weinberger.
\newblock {Cost-Sensitive Tree of Classifiers}.
\newblock In {\em Proceedings of the 30th International Conference on Machine
  Learning (ICML-13)}, pages 133--141, 2013.

\bibitem{zhu2014active}
M.~Zhu, N.~Atanasov, G.~J. Pappas, and K.~Daniilidis.
\newblock {Active deformable part models inference}.
\newblock In {\em Computer Vision--ECCV 2014}, pages 281--296. Springer, 2014.

\end{thebibliography}
}
\end{document}